\documentclass[11pt,a4paper]{article}
\usepackage[hyperref]{emnlp2021}
\usepackage{times}
\usepackage{latexsym}

\usepackage[utf8]{inputenc} 
\usepackage{microtype}
\usepackage{graphicx}
\usepackage{subfigure}
\usepackage{comment}
\usepackage{hyperref}
\usepackage{mathtools}
\usepackage{color}
\usepackage{graphicx}
\usepackage{CJKutf8}
\usepackage{pifont}
\usepackage{url}
\usepackage{enumitem}
\usepackage{multicol,multirow}
\usepackage{arydshln}
\usepackage{makecell}
\usepackage{array}
\usepackage{bm}
\usepackage{cancel}
\usepackage{parskip}
\usepackage[T1]{fontenc}    
\usepackage{booktabs}       
\usepackage{amsmath, amsthm}
\usepackage{amssymb}
\usepackage{amsfonts}
\usepackage[ruled]{algorithm2e}

\usepackage{enumitem}

\newcommand{\sts}{{{\textsc{Seq2Seq}}}\xspace}

\usepackage{microtype}
\definecolor{ao}{rgb}{0.0, 0.5, 0.0}
\definecolor{asparagus}{rgb}{0.53, 0.66, 0.42}
\definecolor{amber}{rgb}{1.0, 0.49, 0.0}
\definecolor{alizarin}{rgb}{0.82, 0.1, 0.26}
\definecolor{applegreen}{rgb}{0.55, 0.71, 0.0}
\definecolor{amethyst}{rgb}{0.6, 0.4, 0.8}
\definecolor{auburn}{rgb}{0.43, 0.21, 0.1}
\def\aclpaperid{1100} 

\title{ConRPG: Paraphrase Generation using Contexts as Regularizer}
\date{}

\author{Yuxian Meng$^\clubsuit$, Xiang Ao$^\blacktriangle$, Qing He$^\blacktriangle$, Xiaofei Sun$^\clubsuit$\\
{\bf Qinghong Han$^\clubsuit$, Fei Wu$^\blacklozenge$, Chun fan$^{\spadesuit\bigstar}$ and Jiwei Li$^{\blacklozenge\clubsuit}$}\\
  $^\blacklozenge$Zhejiang University,
  $^\spadesuit$Computer Center of Peking University,
  $^\bigstar$Peng Cheng Laboratory\\
  $^\blacktriangle$ Key Lab of Intelligent Information Processing of Chinese Academy of Sciences  \\
  $^\clubsuit$Shannon.AI\\
  \{yuxian\_meng,  xiaofei\_sun, qinghong\_han, jiwei\_li\}@shannonai.com\\
  \{aoxiang, heqing\}@ict.ac.cn,
  fanchun@pku.edu.cn,
  wufei@zju.edu.cn
}

\begin{document}
\maketitle

\begin{abstract}
A long-standing issue with paraphrase generation is 
how to obtain
 reliable supervision signals. In this paper, we propose an unsupervised paradigm for paraphrase generation based on the assumption that the probabilities of generating two sentences with the same meaning given the same context should be the same. Inspired by this fundamental idea, we propose a pipelined system which consists of paraphrase candidate generation based on contextual language models, candidate filtering using scoring functions, and paraphrase model training based on  the selected candidates. 

The proposed paradigm offers merits over existing paraphrase generation methods: (1) using the context regularizer on meanings, the model is able to generate massive amounts of high-quality paraphrase pairs; and (2) using human-interpretable scoring functions to select paraphrase pairs from candidates, the proposed framework provides a channel for developers to intervene with the data generation process, leading to a more controllable model. Experimental results across different tasks and datasets demonstrate that the effectiveness of the proposed model  in both supervised and unsupervised setups.\footnote{To appear at EMNLP2021.}

\end{abstract}

\section{Introduction}
Paraphrase generation
\cite{prakash2016neural,cao2016joint,ma-etal-2018-query,wang2018task}
 is the task of generating an output sentence which is semantically identical to a given input sentence but with variations in lexicon or syntax. It is a long-standing problem in the field of natural language processing (NLP) \citep{mckeown-1979-paraphrasing,meteer1988strategies,quirk-etal-2004-monolingual,bannard-callison-burch-2005-paraphrasing,chen2011collecting} and has fundamental applications on end tasks such as semantic parsing \citep{berant2014semantic}, language model pretraining \citep{lewis2020pre} and question answering \citep{dong2017learning}.

A long-standing challenge with paraphrase generation is to obtain
 reliable supervision signals.
One way to resolve this issue is  to manually annotate paraphrase pairs, which is both labor-intensive and expensive. Existing labeled paraphrase datasets \citep{lin2014microsoft,fader-etal-2013-paraphrase,lan-etal-2017-continuously} are 
either of small sizes or  restricted in narrow domains. 
For example, the Quora dataset\footnote{\url{https://www.kaggle.com/c/quora-question-pairs}}  contains 140K paraphrase pairs, 
the size of 
which is insufficient to build a large neural model. 
As another example,     paraphrases in  the larger  MSCOCO \citep{lin2014microsoft} dataset are originally collected as image captions for object recognition, and re-purposed
 for  paraphrase generation.
 The domain for the MSCOCO dataset is thus 
   restricted to captions  depicting visual scenes.


Unsupervised methods, 
such as reinforcement learning \citep{li-etal-2018-paraphrase,Siddique_2020} and 
auto-encoders
 \citep{bowman-etal-2016-generating,roy2019unsupervised}, 
on the other hand, have exhibited their ability for paraphrase generation  in the absence of annotated datasets.  
The core problem with existing unsupervised methods for paraphrase is the lack of an    objective (or reward function in RL) that reliably 
measures the semantic relatedness between two diverse expressions in an unsupervised manner, with which the model can be trained to 
  promote
pairs with the same meaning but diverse expressions. 
For example, 
\citet{hegde2020unsupervised}  crafted unsupervised pseudo training examples by corrupting a sentence and then fed the corrupted one to a pretrained model as the input with the original sentence as the output. 
Since the model is restricted to learning to {\it reconstruct} corrupted sentences, the generated paraphrases tend to be highly similar to the input sentences in terms of both wording and word orders.
The issue in \citet{hegde2020unsupervised}   can be viewed as a microcosm of problems in existing unsupervised methods for paraphrase:
we wish sentences to be diverse in expressions, but do not have a reliable measurement to avoid meaning change when expressions change.  
Additionally, the action of sentence corrupting can be less controllable. 

In this work, we propose 
to address this issue 
by a new paradigm
based
 on the assumption that the probabilities of generating two sentences with the same meaning based on the same context should be the same.
  With this core idea in mind, 
  we propose a pipelined system which consists of  
  the following steps: 
(1) paraphrase candidate generation  by decoding  sentences given its context using a language generation model;
(2) candidate filtering based on scoring functions; and 
(3) paraphrase model training by
training a \sts paraphrase generation model, which can be latter used for supervised finetuning on labeled datasets or  directly used for unsupervised paraphrase generation.

The proposed paradigm offers the following merits over existing methods: (1)
using the context regularizer on meanings, 
 the model is able to generate massive amounts 
 of high-quality paraphrase pairs; 
and (2)  using human-interpretable ranking scores to select paraphrase pairs from candidates, the proposed framework provides a channel for developers to intervene with the data generation process, leading
to a more controllable paraphrase model. 
Extensive experiments across different datasets under both supervised and unsupervised setups demonstrate 
the effectiveness of the proposed model. 

\section{Related Work}  

\paragraph{Supervised Methods} 
for paraphrase generation rely on annotated paraphrase pairs to train the model.
\citet{iyyer2018adversarial,li-etal-2019-decomposable,chen2019controllable,goyal2020neural} leveraged syntactic structures to generate diverse paraphrases with different syntax.
\citet{xu2018d,qian2019exploring} used different semantic embeddings or  generators to produce more diverse paraphrases.
\citet{kazemnejad-etal-2020-paraphrase} proposed a retrieval-based approach to retrieve paraphrase from  a large corpus.
\citet{mallinson2017paraphrasing,sokolov2020neural} casted paraphrase generation as the task of machine translation. \citet{mallinson2017paraphrasing,wieting2017learning} extended the idea of bilingual pivoting for paraphrase generation where the input sentence is first translated into a foreign language, and then translated back as the paraphrase. \citet{sokolov2020neural}  trained a MT model using multilingual parallel data and then finetuned the model using parallel paraphrase data. 

\paragraph{Unsupervised Methods}
\citet{li-etal-2018-paraphrase,Siddique_2020} proposed to generate paraphrases using reinforcement learning, where certain rewarding criteria such as BLEU and ROUGE are optimized.
\citet{bowman-etal-2016-generating,yang2019end} used the generative framework for paraphrase generation by training a variational auto-encoder (VAE) \citep{kingma2013auto} to optimize the lower bound of the reconstruction likelihood for an input sentence. Sentences sampled through the VAE's decoder can be regarded as paraphrases for an input sentence due to the reconstruction optimization target.
\citet{fu2019paraphrase} similarly adopted a generative method but worked at the bag-of-words level.
Other works explored paraphrase generation in an unsupervised manner by using vector quantised VAE (VQ-VAE) \citep{roy2019unsupervised}, simulated annealing \citep{liu2019unsupervised} or disentangled syntactic and semantic spaces \citep{bao-etal-2019-generating}. 
More recently, large-scale language model pretraining has also been proven to benefit paraphrase generation in  both supervised learning \citep{witteveen2019paraphrasing} and unsupervised learning \citep{hegde2020unsupervised}.
\newcite{krishna2020reformulating}
proposed diverse paraphrasing
by warping
the input’s meaning through attribute transfer. 

Regarding soliciting large-scale paraphrase datasets,
\newcite{bannard2005paraphrasing}
used statistical machine translation
methods  
obtain  paraphrases in parallel
text, the technique of which is scaled up by \newcite{ganitkevitch2013ppdb}
to
produce the Paraphrase Database
(PPDB). 
\newcite{wieting2017learning} 
 translate the non-English side of parallel text to obtain paraphrase pairs. 
\newcite{wieting2017paranmt} collected paraphrase dataset with million of pairs via machine translation. 
\newcite{hu2019improved,hu2019parabank} 
produced paraphrases from a bilingual corpus based on the techniques of negative constraints, inference sampling, and clustering. 
A relevant work to ours is \newcite{sun2021sentence}, which harnesses context  to obtain sentence similarity.
\newcite{sun2021sentence} focuses on sentence
similarity rather than paraphrase generation. 

\section{Model}
\begin{figure*}
    \centering
    \includegraphics[scale=0.48]{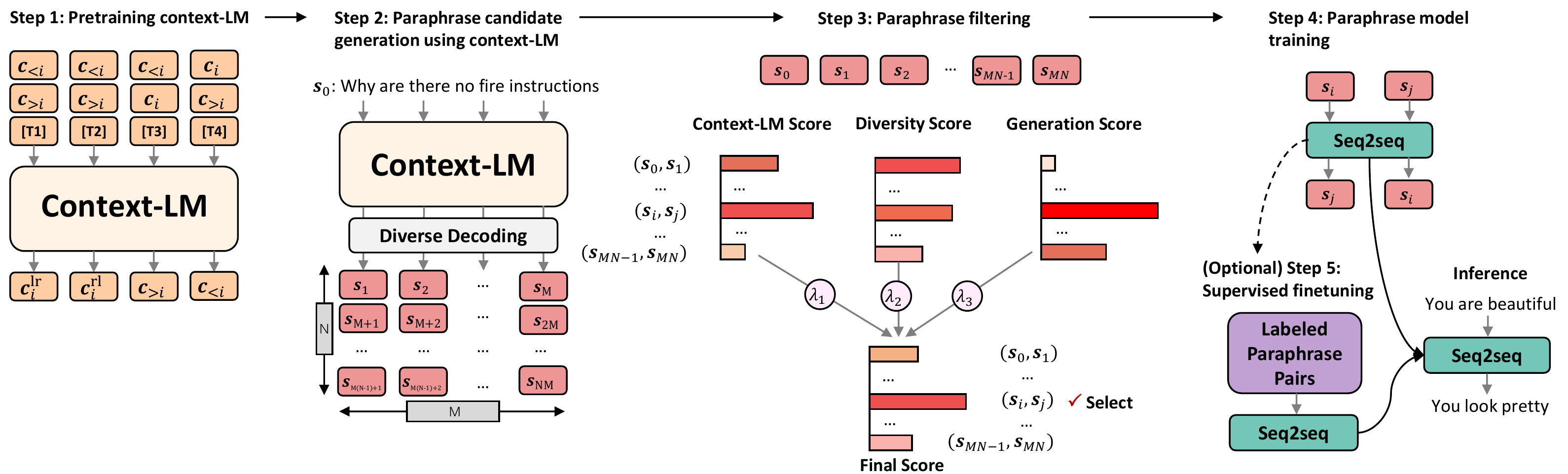}
    \caption{An overview of the proposed ConRPG framework. Step 1: we first train a {\it context-LM} model that predicts the sentence probability in an autoregressive manner given contexts. Step 2: the {\it context-LM} model is used to decode multiple candidate paraphrases with respect to a given context using diverse decoding of beam search. Step 3: paraphrase candidates are
    filtered 
     based on different scoring functions, i.e., the context-LM score, the diversity score and the generation score. Step 4: the selected pair is used to train a \sts model, which can be latter used for supervised finetuning or be directly used for unsupervised paraphrase generation.}
    \label{fig:overview}
\end{figure*}

The key point of the proposed paradigm is to generate paraphrases 
based on the same context. 
This can be done in the following pipelined system: (1) we first train a contextual  language generation model ({\it context-LM}) that predicts sentences given left and right contexts; (2)  
the pretrained contextual generation model  decodes multiple sentences given the same context, and decoded sentences are treated as
paraphrase candidates;
(3) due to the fact that decoded sentences can be extremely noisy, further filtering is needed; 
(4) given the selected paraphrase, a \sts model \citep{sutskever2014sequence} is trained  using one sentence of the paraphrase pair as the source and the other as the target; the \sts model can be directly taken for the use of paraphrase in the unsupervised learning setup, or used as initialization to be further finetuned on  labeled paraphrase datasets in the supervised learning setup.
 An overview of the proposed framework in depicted in Figure \ref{fig:overview},
  the constituent unit of which will be detailed in order below.

\subsection{Training context-LM}
Let $\bm{c}_i=\{w_{i,1}, w_{i,2}, \cdots, w_{i,n}\}$ denote the  $i$-th sentence within the given text, where $n$ is number of words in  $c_j$. $\bm{c}_{i:j}$ denotes the $i$-th to $j$-th sentences. $\bm{c}_{<i}$ and $\bm{c}_{>i}$ respectively denote the preceding and subsequent context of $\bm{c}_i$.
Given  contexts $\bm{c}_{<i}$ and  $\bm{c}_{>i}$, we first train a context-LM by maximizing $p(\bm{c}_i|\bm{c}_{<i},\bm{c}_{>i})$.
The input is a sequence of words and the input representation for each word is the addition of  three embeddings: the sentence-position embedding, token-position embedding and the word embedding. Predicting $\bm{c}_i$ follows a word-by-word fashion.
We consider the style of both left-to-right generation and right-to-left generation to optimize $p(\bm{c}_i|\bm{c}_{<i},\bm{c}_{>i})$, which is respectively given by the following objective:
\begin{equation}
\normalsize
\begin{aligned}
       p(\overrightarrow{\bm{c}}_i|\bm{c}_{<i},\bm{c}_{>i})&=\prod_{j=1}^np(w_{i,j}|\bm{c}_{<i},\bm{c}_{>i},\bm{w}_{i,<j})\\ p(\overleftarrow{\bm{c}}_i|\bm{c}_{<i},\bm{c}_{>i})&=\prod_{j=n}^1p(w_{i,j}|\bm{c}_{<i},\bm{c}_{>i},\bm{w}_{i,>j})
    \end{aligned}
\end{equation}
$p(\bm{c}_i|\bm{c}_{<i},\bm{c}_{>i})$ models the forward probability from contexts to sentences. 
For two sentences of the same meaning, the probability of generating contexts given the two sentences should be also the same, which correspond to the backward probability given 
from sentences  to contexts. 
This is akin to the 
bi-directional mutual-information based generation strategy
 \citep{fang2015captions,jiwei2016diversity,li2016mutual,wang2021modeling}. 
 The backward probability can be modeled by 
  predicting preceding contexts given subsequent contexts $p(\bm{c}_{<i}|\bm{c}_i,\bm{c}_{>i})$ and to predict subsequent contexts given preceding contexts $p(\bm{c}_{>i}|\bm{c}_{<i},\bm{c}_i)$. 
  
We implement the above models, i.e. $p(\overrightarrow{\bm{c}}_i|\bm{c}_{<i},\bm{c}_{>i})$, $p(\overleftarrow{\bm{c}}_i|\bm{c}_{<i},\bm{c}_{>i})$, $p(\bm{c}_{<i}| \bm{c}_i,\bm{c}_{>i})$,
  $p(\bm{c}_{>i}|\bm{c}_{<i},\bm{c}_i)$
based on the \sts structure on a subset of CommonCrawl containing 10 billion tokens in total.
We use Transformers as the backbone \cite{vaswani2017attention}\footnote{
The four models share the same structure
 but with a special objective-specific token appended to the model input notifying different objectives.}
with the number of encoder blocks, decoder blocks, 
the number of heads,
$d_{model}$ and $d_ff$
 set to 6, 6, 8, 512 and 2048. 
We use adam  \citep{kingma2014adam}  for optimization,
with learning rate of 1e-4, $\beta_1$ = 0.9,
$\beta_2$ = 0.999.
We consider a maximum number of +800 and -800 tokens as contexts. 

\subsection{Paraphrase Candidate Generation}
Using the pretrained {\it context-LM} models, we generate potential paraphrases by decoding multiple outputs given the input sentence only based on $p(\overrightarrow{\bm{c}}_i|\bm{c}_{<i},\bm{c}_{>i})$. 
The other three contextual objectives, i.e., $p(\overleftarrow{\bm{c}}_i|\bm{c}_{<i},\bm{c}_{>i})$, $p(\bm{c}_{<i}|\bm{c}_i,\bm{c}_{>i})$ and
  $p(\bm{c}_{>i}|\bm{c}_{<i},\bm{c}_i)$ cannot be readily used at the decoding stage since their computations
 require the completion of the target generation. 
They will thus be used at the later reranking stage. 
We use diverse decoding strategy of beam search \citep{li2016simple} to generate diverse candidates. 
Decoded candidates are guaranteed to be fluent.\footnote{Implementation-wise, we first cache all the possible candidate paraphrase pairs for all input context sentences. These pairs are then used for filtering, as will be detailed in the next section. We also impose a constraint that at most one paraphrase pair with respect to an input context is selected for training the final \sts model (Section \ref{sec:training}).}

\subsection{Paraphrase Filtering}
The decoded andidates can not be readily used since (1)  
 candidates often differ
only by punctuation or minor morphological variations, with almost all words overlapping, and  (2) many of them are not of the same meaning. 
We thus propose to further rank a candidate pairs. The ranking model consists of three parts:
\subsubsection{Context LM Score} 
For a pair of sentences 
$\bm{s}_1$ and $\bm{s}_2$
of the same meaning, differences between  
the probabilities of generating them given the same context should be very similar. 
In the same way, the probabilities of predicting left and right contexts given the two sentences with the same meaning should also be  similar.
The ranking scoring function to rank $(\bm{s}_1, \bm{s}_2)$ consists the following parts:
(1) the probability difference in generating two sentences given contexts, i.e., $ \frac{1}{|\bm{s}|} \log p(\overrightarrow{\bm{s}}| \bm{c}_{<i}, \bm{c}_{>i})$
and $ \frac{1}{|\bm{s}|} \log p(\overleftarrow{\bm{s}}| \bm{c}_{<i}, \bm{c}_{>i})$;
(2) 
the probability difference in generating contexts given two sentences, i.e., $\frac{1}{|\bm{c}_{<i}|} |\log p(\bm{c}_{<i}| \bm{s}, \bm{c}_{>i})$ and 
$\frac{1}{|\bm{c}_{<i}|} |\log p(\bm{c}_{<i}| \bm{s}, \bm{c}_{>i})$. 
\subsubsection{Lexicon and Syntactic Diversity }
Two identical sentences will have the optimal score, which does not serve our purpose since we wish paraphrases to be as diverse as possible \cite{li-etal-2018-paraphrase}.  
We consider two types of diversity:
(1) lexicon diversity, which encourages individual word or phrase replacements using synonyms; and 
(2) syntactic diversity, which encourages syntactic shifting such as heavy NP shift. 
Lexicon diversity
is measured by the unigram-based Jaccard distance between two sentences. 
Syntactic diversity is measured by  the relative position change for shared unigrams. If $\bm{s}_2$  contains multiple copies of a word $w$ in $\bm{s}_1$, we pick the nearest copy.  Let $\text{pos}_{\bm{s}}(w)$ denote the position index of $w$ in $\bm{s}$. 
The combination of
lexicon and syntactic diversity  is given as follows:
\begin{equation}
\begin{aligned}
&\text{S}_{\text{diversity}}(\bm{s}_1, \bm{s}_2) = \beta_1\frac{|\bm{s}_1\cap \bm{s}_2|}{|\bm{s}_1\cup \bm{s}_2|}  \\+
&\beta_2\frac{1}{|\bm{s}_1\cap \bm{s}_2|}\sum_{w\in \bm{s}_1\cap \bm{s}_2} \frac{|\text{pos}_{\bm{s}_1}(w)-\text{pos}_{\bm{s}_2}(w)|}{\text{max}(|\bm{s}_1|,|\bm{s}_2|)}
\end{aligned}
\end{equation}
where the first part denotes the unigram Jaccard distance, and the second part denotes the relative position change for unigrams. 

\subsubsection{Mutual Generation Score}
It is noteworthy that an intrinsic drawback of the proposed methodology (and other paraphrase generation methods as well) is that, 
two sentences that can fit into the same context are not necessarily of the exactly same meaning, e,g,  
sentences with very similar general semantics but vary in some specific details (e.g., number).
Think about two sentences,  {\it I spent 5 dollars on this mug.} v.s.  
{\it I spent 6 dollars on this mug}. 
If one sentence fits  into certain contexts, it is very likely that the other sentence  will also fit in. 
The issue can be alleviated with more contexts considered, but the practical problem still remains because our model can only consider a very limited number of contexts due to 
hardware limitations.

We propose a strategy to address this drawback. 
The strategy is  inspired by the famous idiom that 
 ``{\it Happy families are all alike; every unhappy family is unhappy in its own way}".
 Paraphrases share the same meaning in the vector space, and there should be a direct and easy mapping between them.  
 Non-paraphrases are different in random ways.
 It is thus easier to predict a paraphrase given a sentence than predict a specific non-paraphrase given the sentence. 
For example,  
  $p(\text{``six dollars''} | \text{``6 dollars''})$
should be higher than generating a random sentence give the sentence
e.g., $p(\text{``5 dollars''} | \text{``6 dollars''})$. 
This is because, there are so many ways to generate  non-paraphrase 
e.g., $p(\text{``5 dollars''} | \text{``6 dollars''})$ and $p(\text{``7 dollars''} | \text{``6 dollars''})$, etc. 
These non-paraphrases split the probability, making the probability for an individual non-paraphrase  low. 
To this end, we train a \sts model \citep{sutskever2014sequence} on 
8 million
 pairs of decoded candidates using Transformer-based. 
Next, using this  model, we give the mutual decoding score for any sentence pair $(\bm{s}_1, \bm{s}_2)$ as follows:
\begin{equation}
\text{S}_{\text{generation}}= \gamma_1\frac{1}{|\bm{s}_1|}\log p(\bm{s}_1|\bm{s}_2)+\gamma_2\frac{1}{|\bm{s}_2|}\log p(\bm{s}_2|\bm{s}_1)
\label{gen}
\end{equation}
For  a sentence pair of the same meaning, they should have higher values of Eq.\ref{gen}.

\subsubsection{Final Ranking Model}
\label{ranking}
The final ranking score
 is a linear combination of scores above as follows:
\begin{equation}
\begin{aligned}
&\text{S}(\bm{s}_1, \bm{s}_2) =  \text{S}_{\text{context}}(\bm{s}_1, \bm{s}_2)  \\
&+ \text{S}_{\text{diversity}}(\bm{s}_1, \bm{s}_2)  + \text{S}_{\text{generation}}(\bm{s}_1, \bm{s}_2) 
\end{aligned}
\end{equation}
We build a ranking model to  learn
weights (i.e., $\alpha$, $\beta$, $\gamma$, eight parameters in total). To train the ranking  model, 
we annotate a small proportion of data  on Amazon Mechanical Turk. 
A Turker is first given a sentence (denoted by $a$) randomly picked from 
the
 candidate pool. Next, the Turker is given two other decoded sentences ($b_1$ and $b_2$), and is asked to decide which one is a better paraphrase of $a$,
in terms of three aspects: 
(1) semantics: whether the two sentences are of the same semantic meaning;
(2) diversity: whether the two sentences are diverse in expressions; and 
(3) fluency: whether the generated paraphrase is fluent. 
Ties are allowed and will be further removed. We  labeled a total number of  2K pairs.  
Let $b_{+}$ denote the better paraphrase by annotators, and $b_{-}$ denote the other. 
 Based on the labeled dataset,  a simple pairwise ranking model \cite{liu2011learning}  is built for weight learning:
\begin{equation}
L = \max(0, 1+S(a, b_{+})-S(a, b_{-}))
\end{equation}
It is worth noting that the filtering module 
 provides a channel for developers to intervene with the data generation process, as  developers can develop their own scoring functions to generate paraphrases of specific features. This leads to a more controllable paraphrase model. 
 
\subsection{Paraphrase Model Training}
\label{sec:training}
We select 10 million paraphrase pairs in total based on criteria above, on which 
we train a \sts model for paraphrase generation, using one sentence of the pair as the input, and the other as the output. 
We use the  Transformer-base \citep{vaswani2017attention} as the model backbone.
We use Adam \citep{kingma2014adam} with learning rate of 1e-4, $\beta_1=0.9$,
$\beta_2=0.98$ and a warmup step of 4K.
The trained model can be directly used for paraphrase generation in the unsupervised setup \citep{roy2019unsupervised,liu2019unsupervised}.

 For the supervised setup  \cite{witteveen2019paraphrasing,kazemnejad-etal-2020-paraphrase,hegde2020unsupervised}, where we have 
 pairs of paraphrases containing
  sources from a source domain and paraphrases of sources from a target domain,
  we can fine-tune the pretrained model on the supervised paraphrase pairs, where we initialize the model using the 
pre-trained
 model, and run additional iterations on the supervised dataset. 
Again, we use adam  \citep{kingma2014adam} for fine-tuning, with $\beta_1=0.9$,
$\beta_2=0.98$. Batch size, learning rate 
and the number of iterations
are treated as hyper-parameters, to be tuned on the dev set.


It is worth nothing that the \sts model here is different from the \sts model in the filtering stage, as the model here is
trained on the remaining paraphrase pairs and 
 used for direct paraphrase generation, while the other is trained on the noisy pairs and used for candidate filtering. 

\section{Experiments}
\subsection{Datasets}
We carry out experiments in both supervised and unsupervised setups. 
For the unsupervised setting, we use the Quora, Wikianswers \citep{fader-etal-2013-paraphrase}, MSCOCO \citep{lin2014microsoft} and Twitter  \citep{lan-etal-2017-continuously} datasets. For the supervised setting, we use the Quora and Wikianswers datasets. 
\begin{itemize}[noitemsep]
  \item {\bf Quora}: The Quora question pair dataset\footnote{\url{https://www.kaggle.com/c/quora-question-pairs}} contains 140K parallel paraphrases and 260K non-parallel sentences. We follow the standard setup in \citet{miao2019cgmh} where 3K and 30K paraphrase pairs are respectively used for validation and test. 
  \item {\bf Wikianswers}: The Wikianswers dataset \citep{fader-etal-2013-paraphrase} contains 2.3M paraphrase pairs extracted from the Wikianswers website. We follow \citet{liu2019unsupervised} to randomly pick 5K pairs for validation and 20K for test.\footnote{Note that the selected data is different from \citet{liu2019unsupervised} but is comparable in the statistical sense.} 
  \item {\bf MSCOCO}: The MSCOCO dataset \citep{lin2014microsoft} contains over 500K paraphrase pairs for 120K image captions. We follow the standard dataset split and the evaluation protocol in \citet{liu2019unsupervised}.
  \item {\bf Twitter}: The Twitter dataset is collected via linked tweets through shared URLs \citep{lan-etal-2017-continuously}, which originally contains 50K paraphrase pairs. We follow the data split in \citet{liu2019unsupervised}.
\end{itemize}

\subsection{Baselines and Metrics}
We compare our proposed ConRPG model to the following existing  paraphrase generation models.
Unsupervised paraphrase generation baselines we consider include:
\begin{itemize}[noitemsep]
  \item {\bf VAE}:  
  paraphrases are sampled by encoding a sentence to a continuous space using  (VAEs) \cite{bowman-etal-2016-generating}. 
  \item {\bf Lag VAE}: 
  A sophisticated version of VAE to deal with the posterior collapse issue \citet{he2019lagging}.
  \item {\bf CGMH}: \citet{miao2019cgmh} used Metropolis–Hastings sampling for constrained sentence generation, where a word can be deleted, replaced or inserted into the current sentence based on the sampling distribution.
  \item {\bf UPSA}: \citet{liu2019unsupervised} proposed to treat unsupervised paraphrase generation as an optimization problem with an objective combining semantic similarity, expression diversity and language fluency being optimized  using simulated annealing.
  \item {\bf Corruption}: \cite{hegde2020unsupervised} proposed strategy of corrupting input sentences by  removing stop words and randomly shuffle and replace the remaining 20\% words. We use BART \cite{lewis2019bart} as the backbone to generate targets given corrupted inputs.  
\end{itemize}
Results for 
VAE, Lag VAE, CGMH and UPSA on different datasets are copied from \newcite{miao2019cgmh} and \newcite{liu2019unsupervised}. 
Supervised paraphrase generation baselines include:
\begin{itemize}[noitemsep]
  \item {\bf ResidualLSTM}: \citet{prakash-etal-2016-neural} deepened the LSTM network by stacking multiple layers with residual connection.
  \item {\bf VAE-SVG-eq}: \citet{gupta2018deep} combined VAEs with LSTMs for paraphrase generation. Both encoder and decoder are conditioned on the source input sentence so that more consistent paraphrases can be generated.
  \item {\bf Pointer}: \citet{see-etal-2017-get} augmented the standard \sts model by using a pointer mechanism which can copy source words in the input rather than decode from scratch.
  \item {\bf Transformer}: \citet{vaswani2017attention} proposed the Transformer architecture which is based on the self-attention mechanism.
  \item {\bf DNPG}: \citet{li-etal-2019-decomposable} proposed a Transformer-based model that can learn and generate paraphrases at different  granularities.
\end{itemize}
Results for 
 ResidualLSTM, VAE-SVG-eq, Pointer, Transformer on various datasets are copied from \newcite{li-etal-2019-decomposable}. 
For reference purposes, we also implement the {\bf BT} baseline inspired by the idea of back-translation \citep{sennrich2016back-translation,wieting2017learning}. 
We use Transformer-large as the backbone. BT is trained end-to-end on WMT'14 En$\leftrightarrow$Fr.\footnote{\newcite{wieting2017learning,wieting2017paranmt} suggested little difference among Czech,
German, and French as source languages for backtranslation. We use En$\leftrightarrow$Fr since it contains more parallel data than other language pairs. } 
A paraphrase pair is obtained by pairing the English sentence  in the original dataset and the translation of  the French sentence. 
Next we train a Transformer-large model on paraphrase pairs. 

We evaluate all models using BLEU \citep{papineni2002bleu}, iBLEU \citep{sun2012joint} and ROUGE scores \citep{lin-2004-rouge} . The iBLEU score penalizes the similarity of the generated paraphrase with respect to the original input sentence. Concretely, the iBLEU score of a triple of sentences $(\bm{s},\bm{r},\bm{c})$ is given by:
\begin{equation}
\begin{aligned}
   \text{iBLEU}(\bm{s},\bm{r},\bm{c})&=\alpha\text{BLEU}(\bm{c},\bm{r})\\
   &-(1-\alpha)\text{BLEU}(\bm{c},\bm{s}) 
   \end{aligned}
\end{equation}
where $\bm{s}$ is the input sentence, $\bm{r}$ is the reference paraphrase and $\bm{c}$ is generated paraphrase.  $\alpha$ is set to 0.8 following prior works.

\subsection{In-domain Results}
We first show the in-domain results in Table \ref{tab:in-domain}. As can be seen, across all datasets, the proposed ConRPG model significantly outperforms baselines in both supervised and unsupervised settings. For the supervised setting, ConRPG yields an approximately   2-point gain
across different evaluation metrics 
 against the strong DNPG baseline on both Quora and Wikianswers. We also observe that the BT model 
 is able to achieve competitive results. This shows that back-translation can serve as a simple yet strong baseline for paragraph generation. For the unsupervised setting, we observe substantial performance boosts 
 brought by ConRPG  
 over existing unsupervised methods including the state-of-the-art model UPSA. It is also surprising to see that 
 unsupervised 
 ConRPG  outperforms the supervised VAE-SVG-eq model and achieves comparable results to 
  supervised baselines such as
 Transformer.

\begin{table}[!t]
  \centering
  \small
  \scalebox{0.8}{
  \begin{tabular}{clcccc}\toprule
    & {\bf Model}  & {\bf iBLEU} & {\bf BLEU} & {\bf R1} & {\bf R2}\\\midrule
    \multirow{18}{*}{\rotatebox{90}{{Supervised}}} & \multicolumn{5}{c}{\underline{\it Quora}}\vspace{1pt}\\
    & {\it ResidualLSTM} & 12.67  & 17.57 &  59.22 & 32.40\\
    & {\it VAE-SVG-eq} &15.17 & 20.04 & 59.98 & 33.30\\
    & {\it Pointer} &  16.79 & 22.65 & 61.96 & 36.07\\
    & {\it Transformer} & 16.25 & 21.73 & 60.25 & 33.45\\
    & {\it Transformer+Copy} &  17.98 & 24.77 & 63.34 & 37.31\\
    & {\it DNPG} & 18.01 & 25.03 & 63.73 & 37.75\\
    & {\it BT} &  17.73 & 24.99 & 62.07 & 36.12 \\
    & {\it ConRPG} & {\bf 19.96} &{\bf 26.81} & {\bf 65.03} & {\bf 38.49}\\
    \specialrule{0em}{1pt}{1pt}
    \cdashline{2-6}
    \specialrule{0em}{1pt}{1pt}
    & \multicolumn{5}{c}{\underline{\it Wikianswers}}\vspace{1pt}\\
    & {\it ResidualLSTM} & 22.94& 27.36 &48.52& 18.71    \\
    & {\it VAE-SVG-eq} &26.35 &32.98 &50.93 &19.11\\
    & {\it Pointer} & 31.98 &39.36 &57.19 &25.38\\
    & {\it Transformer} & 27.70& 33.01& 51.85& 20.70\\
    & {\it Transformer+Copy} &  31.43 &37.88 &55.88 &23.37\\
    & {\it DNPG} & 34.15& 41.64& 57.32& 25.88\\
    & {\it BT} &{33.65}&39.70 & 56.89 & 25.22 \\
    & {\it ConRPG} &\bf{35.28} & {\bf 42.25} & {\bf  58.40} & {\bf 26.44}\\\cmidrule{1-6} 
    \multirow{28}{*}{\rotatebox{90}{{Unsupervised}}} & \multicolumn{5}{c}{\underline{\it Quora}}\vspace{1pt}\\
    & {\it VAE} & 8.16& 13.96 &44.55 &22.64 \\
    & {\it Lag VAE} & 8.73& 15.52 &49.20 &26.07 \\
    & {\it CGMH} & 9.94& 15.73& 48.73 &26.12 \\
    & {\it UPSA} & 12.03 &18.21 &59.51 &32.63  \\
    & {\it BT} & 11.64& 11.59 & 58.20 & 32.04 \\
        & {\it Corruption} & 12.32 & 17.97 & 59.14 & 32.44 \\
    & {\it ConRPG} & {\bf 12.68} &{\bf  18.31} & {\bf  59.62} & {\bf 33.10}\\
    \specialrule{0em}{1pt}{1pt}
    \cdashline{2-6}
    \specialrule{0em}{1pt}{1pt}
    & \multicolumn{5}{c}{\underline{\it Wikianswers}}\vspace{1pt}\\
    & {\it VAE} &  17.92 &24.13 &31.87 &12.08 \\
    & {\it Lag VAE} & 18.38& 25.08 &35.65 &13.21 \\
    & {\it CGMH} &  20.05 &26.45 &43.31 &16.53 \\
    & {\it UPSA} & 24.84& 32.39 &54.12 &21.45\\
    & {\it BT} & 24.17 & 31.75&  53.69& 20.63\\
     & {\it Corruption} &  24.40 & 32.05 & 53.77 & 21.22 \\
    & {\it ConRPG} &{\bf  25.98} & {\bf 32.89} & {\bf 54.65} &{\bf  22.25}\\
    \specialrule{0em}{1pt}{1pt}
    \cdashline{2-6}
    \specialrule{0em}{1pt}{1pt}
    & \multicolumn{5}{c}{\underline{\it MSCOCO}}\vspace{1pt}\\
    & {\it VAE} & 7.48 &11.09& 31.78& 8.66 \\
    & {\it Lag VAE} &  7.69 &11.63 &32.20 &8.71 \\
    & {\it CGMH} & 7.84& 11.45& 32.19 &8.67 \\
    & {\it UPSA} & 9.26 &14.16 &37.18& 11.21 \\
    & {\it BT} & 9.72 & 14.36 &37.64  &11.81 \\
    & {\it Corruption} & 10.32 & 15.60 & 38.12 & 12.40 \\
    & {\it ConRPG} & {\bf 11.17} & {\bf 16.98} & {\bf 39.42} & {\bf 13.50}\\
    \specialrule{0em}{1pt}{1pt}
    \cdashline{2-6}
    \specialrule{0em}{1pt}{1pt}
    & \multicolumn{5}{c}{\underline{\it Twitter}}\vspace{1pt}\\
    & {\it VAE} & 2.92& 3.46& 15.13& 3.40    \\
    & {\it Lag VAE} & 3.15& 3.74& 17.20 &3.79 \\
    & {\it CGMH} & 4.18& 5.32& 19.96 &5.44\\
    & {\it UPSA} &  4.93& 6.87 &28.34 &8.53 \\
    & {\it BT} &5.11  & 6.99 &29.11  &8.95 \\
     & {\it Corruption} & 5.32 & 7.11 & 29.80 & 9.32 \\
    & {\it ConRPG} & {\bf 5.83} & {\bf 7.32} & {\bf 30.81} & {\bf 10.08}\\\bottomrule
  \end{tabular}
  }
  \caption{In-domain performances of different models for both supervised and unsupervised setups. }
  \label{tab:in-domain}
\end{table}
\subsection{Domain-adapted Results}
We test the domain adaptation ability of the proposed method on the Quora and Wikianswers datasets.
Results are shown in Table \ref{tab:cross-domain}. We can see that ConRPG significantly outperforms baselines in both settings, i.e. {\it Quora$\rightarrow$Wikianswers} and {\it Wikianswers$\rightarrow$Quora}, showing the better ability of ConRPG for domain adaptation. 
\subsection{Human Evaluation}
To further validate the performance of the proposed model, we sample 400  
sentences from the Quora test set for human evaluation.
We assign the input sentence and its generated paraphrase to  three human annotators at 
Amazon Mechanical Turk (AMT), with “> 95$\%$ HIT approval rate”.
Turkers are asked to  evaluate the  quality of
generated paraphrases by considering three aspects 
semantics, diversity and fluency, as detailed in Section \ref{ranking}. 
Each paraphrase is labeled 
 by a
5-point scale (Strongly Agree, Agree, Unsure, Disagree, Strongly Disagree) and assigned to three annotators.
We evaluate three models: 
BT,  Corruption, and the proposed ConRPG model. 
The  Cohen's kappa score \citep{mchugh2012interrater} for the three aspects are 0.55, 0.52 and 0.49, indicating moderate inter-annotator agreement. 
Table \ref{tab:human} presents the human evaluation results. As can be seen from the table, the proposed ConRPG model significantly outperforms BT and Corruption in terms of all three aspects, which is consistent with the automatic evaluation results. 

\begin{table}[t]
    \centering
    \small
    \scalebox{0.85}{
    \begin{tabular}{lccc}\toprule
    {\bf Model} & {\bf Semantics} & {\bf Diversity} & {\bf Fluency}\\\midrule
    {\it ConRPG} & {\bf 3.78 (0.5)} & {\bf 4.01 (0.4)} &{\bf 4.21 (0.3)} \\
    {\it Corruption} & 3.14 (0.6) &  3.17 (0.5) & 4.19 (0.4) \\
    {\it BT} & 3.04 (0.6) & 3.32 (0.5) & 3.89 (0.4)\\\bottomrule
    \end{tabular}
    }
    \caption{Human evaluation results for BT, UPSA and ConRPG under the unsupervised setup.}
    \label{tab:human}
\end{table}

\begin{table}[!t]
  \centering
  \small
  \scalebox{0.85}{
  \begin{tabular}{lcccc}\toprule
      {\bf Model} & {\bf iBLEU} & {\bf BLEU} & {\bf R1} & {\bf R2}\\\midrule 
      \multicolumn{5}{c}{\underline{\it Wikianswers$\rightarrow$Quora}}\\
        {\it Pointer}& 5.04 &6.96& 41.89 &12.77 \\
        {\it Transformer+Copy} &  6.17& 8.15& 44.89 &14.79 \\
        {\it DNPG} &10.39 &16.98 &56.01& 28.61\\
        {\it {BT}} & 12.54&17.98 &59.43 &32.54 \\
        {\it ConRPG} &{\bf 13.25} & {\bf 19.28} &{\bf 60.55}&  {\bf 34.17}\\
        \specialrule{0em}{1pt}{1pt}
        \cdashline{1-5}
        \specialrule{0em}{1pt}{1pt}
    \multicolumn{5}{c}{\underline{\it Quora$\rightarrow$Wikianswers}}\\
        {\it Pointer}  &21.87 &27.94& 53.99 &20.85 \\
        {\it Transformer+Copy}&  23.25& 29.22& 53.33& 21.02 \\
        {\it DNPG}  & 25.60 & 35.12 &56.17 &23.65 \\
        {\it {BT}} &26.11 &35.28 &57.29 & 23.88\\
        {\it ConRPG} &{\bf 28.14}  &{\bf 37.93}  &{\bf 57.98} & {\bf 25.32}\\\bottomrule
  \end{tabular}
  }
  \caption{Domain-adapted performances.}
  \label{tab:cross-domain}
\end{table}
\section{Ablation Study}
\subsection{Size of Data to Train {\it context-LM}}
First, we would like to understand how the data size for training {\it context-LM} effects the downstream performance of Wikianswers.
Table \ref{tab:size} presents the results where the training data size is respectively 10M, 100M, 1B and 10B tokens.
We can observe that with more training data, downstream performances under both setups increase. This is because more training data leads to a more reliable context regularization, and thus the trained model can produce paraphrases with higher qualities.

\begin{table*}
  \centering
  \small
  \scalebox{0.9}{
  \begin{tabular}{p{5.5cm}p{5.5cm}p{5.5cm}}
  \toprule
  {\bf Input} & {\bf Corruption} & {\bf ConRPG} \\
   \midrule
   What should be the first computer table language I learn?& What should be my first computer table language? & 
   If I want to learn a programming language, which one should I learn first?
   \\\cline{1-3}
   How do I overcome my shyness with women? & How do I overcome  shyness with women?&
   How can I overcome being shy when women are around. 
    \\\cline{1-3}
   Should Harry Potter have ended up with Cho Chang?& Should Harry Potter be with Chang Cho?&
   Should Harry Potter and Cho Chang end up being together? \\\cline{1-3}
   How do I become a data scientist in Malaysia?
 &  How can I become a  scientist in Malaysia?
 &  What do I need to do if I want to become a data scientist in Malaysia?   \\\cline{1-3}
 What are the rate common regrets in old age? &What are the  common regrets in old age?
  & What do old people regret the most?\\\cline{1-3}
   \bottomrule
  \end{tabular}
  }
  \caption{Sampled paraphrases from the Corruption model and  the ConRPG model.}
  \label{tab:example}
\end{table*}

\begin{table}
    \centering
    \small
    \scalebox{0.9}{
    \begin{tabular}{lcccc}\toprule
    {\bf Setup} & {\bf 10M} & {\bf 100M} & {\bf 1B} & {\bf 10B} \\\midrule 
    {\it Unsupervised} & 13.8 & 19.2 & 24.9 & 26.0 \\
    {\it Supervised} & 31.4 & 32.5 & 34.4 & 35.3  \\\bottomrule
    \end{tabular}
    }
    \caption{The effect of data size for training {\it context-LM}. The iBLEU score is reported on Wikianswers.}
    \label{tab:size}
\end{table}

\begin{table}
    \centering
    \small
    \scalebox{0.9}{
    \begin{tabular}{lccccc}\toprule
    {\bf Setup} &  {\bf 100} & {\bf 400} & {\bf 800}\\\midrule 
    {\it Unsupervised} & 21.9 & 25.1  & { 26.0} \\
    {\it Supervised}  & 31.6 & 34.1 & { 35.3} \\\bottomrule
    \end{tabular}}
    \caption{The effect of context length for  {\it context-LM}. The iBLEU score is reported on Wikianswers.}
    \label{tab:context-length}
\end{table}
\subsection{Context Length to Train {\it context-LM}}
Table \ref{tab:context-length} presents the influence of context length used to train  {\it context-LM} on Wikianswers.
As can be seen, the performance is sensitive to the context length, which can be explained by the fact that more contexts lead to a significantly better language modeling. 

\subsection{Percentage of Selected Paraphrase Pairs}
Table \ref{tab:candidate}
presents
 the impact of  the percentage of selected paraphrase pairs in the filtering process on the final performance of Wikianswers. 
 We tune the ratio $\rho$, which is defined as the number of remaining paraphrase pairs  divided by the number of input contexts for {\it context-LM}. $\rho=1$ is what we use in this work: selecting the top-1 paraphrase pair for each input context makes the number of remaining pairs equal to the number of input contexts. 
As expected, either too few or too many selected paraphrase pairs  leads to worse performances.  Too few pairs lead to  insufficient training and too many pairs lead to noise that harm the final performance. 
A tricky balance of the percentage of selected paraphrase pairs is thus crucial for better final performances.

\begin{table}[t]
    \centering
    \small
    \scalebox{0.9}{
    \begin{tabular}{lcccc}\toprule
    {\bf Setup/iBLEU} & {\bf $\rho$=0.01} & {\bf $\rho$=0.1} & {\bf $\rho$=1} & {\bf $\rho$=5}\\\midrule 
    {\it Unsupervised} & 20.8 &25.0&  { 26.0}  &24.2 \\
    {\it Supervised} & 32.5 &  34.8  &{ 35.3} & 34.8 \\\bottomrule
    \end{tabular}
    }
    \caption{The effect of percentage of selected candidates for candidate reranking on Wikianswers.}
    \label{tab:candidate}
\end{table} 


\subsection{Effects of Different Modules}
We are interested in the effectiveness of each module within the proposed framework. 
Table \ref{tab:module} shows the performance:  

(1) Removing the entire {\it filtering} module leads to the most degradation in performance, which is in line with our expectation: with filtering, high quality paraphrase pairs that both share the same meaning and are diverse in lexicon can be selected for training the final paraphrase generation model.

(2) Removing {\it backward}, i.e., $p(\bm{c}_{<i}|\bm{c}_i,\bm{c}_{>i})$  and  $p(\bm{c}_{>i}|\bm{c}_{<i},\bm{c}_i)$ , leads to the second largest performance reduction. This is because removing {\it backward} greatly weakens the strength of context regularization, introducing more noise for the subsequent paraphrase filtering phase.

(3) Removing {\it right-to-left}, i.e., $p(\overleftarrow{\bm{c}}_i|\bm{c}_{<i},\bm{c}_{>i})$, leads to a slight drop in performance.

(4) Removing the diversity score or the generation score  harms model performances. This observation verifies that using scores from different aspects significantly helps 
 paraphrase quality.

\begin{table}[t]
    \centering
    \small
    \scalebox{0.9}{
    \begin{tabular}{ll}\toprule
    {\bf Model} & {\bf iBLEU}   \\\midrule 
    {\it Full}       &  25.98  \\
    {\it w/o filtering} & 20.12 (-5.86) \\
    {\it w/o backward}&  23.29 (-2.69)\\
    {\it w/o right-to-left}  &  25.56 (-0.42)\\
    {\it w/o diversity} & 23.99 (-1.99)\\
    {\it w/o generation} &24.80 (-1.18)\\
    \bottomrule
    \end{tabular}
    }
    \caption{The effect of different modules within ConRPG. {\it w/o filtering} means removing the filtering phase and randomly choosing a paraphrase pair for each context. {\it w/o backward} means removing the $p(\bm{c}_{<i}|\bm{c}_i,\bm{c}_{>i})$  and  $p(\bm{c}_{>i}|\bm{c}_{<i},\bm{c}_i)$ training objectives. {\it w/o right-to-left} means removing the $p(\overleftarrow{\bm{c}}_i|\bm{c}_{<i},\bm{c}_{>i})$ training objective. {\it w/o diversity} means removing the diversity score and {\it w/o generation} means removing the generation score.}
    \label{tab:module}
\end{table}   

\section{Conclusion}
In this paper, we propose  ConRPG, a  paradigm for paraphrase generation using context regularizer. ConRPG is 
based on the assumption that the probabilities of generating two sentences with the same meaning based on the same context should be the same.
We acknowledge that the current system is rather complicated, which requires multiple pipelines and modules  to  build.
We will simplify the system in future work.

\bibliography{emnlp2020}
\bibliographystyle{acl_natbib}

\end{document}